# CMCNet: Aligning Ultrasound Image Embeddings with Textual TI-RADS Representations for Fine-Grained Thyroid Classification

Bingxin Yu[1], Xueli Wang[2,3], Jerry Zhou[2,3,4], Wenyan Wang[2,3], Li Wen[5], Lan Huang[2,3], Xin Feng[6], Fengfeng Zhou[2,3,*], Kewei Li[2,3,*].

1 China-Japan Union Hospital of Jilin University, Changchun, 130012, P.R. China.

2 College of Computer Science and Technology, Jilin University, Changchun, 130012, P.R. China.

3 Key Laboratory of Symbolic Computation and Knowledge Engineering of Ministry of Education, Jilin University, Changchun, 130012, P.R. China.

4 Shanghai Pinghe School, 261 Huang Yang Road, Pudong, Shanghai, 201206, P.R. China.

5 Center for Biomarker Discovery and Validation, Institute of Clinical Medicine, Peking Union Medical College Hospital, Chinese Academy of Medical Science, Beijing 100073, P.R. China

6 School of Science, Jilin University of Chemical Technology, Jilin 130000, P.R. China.

* Correspondence may be addressed to F.Z. (FengfengZhou@gmail.com) or K.L. (kwbb1997@gmail.com).

## Abstract

Ultrasound is the primary imaging modality for assessing thyroid nodules, and the ACR TI-RADS framework standardizes diagnosis through five ultrasound feature categories that are aggregated into five risk levels (TR1–TR5). Although widely adopted in clinical practice, most deep learning approaches focus on binary malignancy classification, while multi-class prediction and explicit utilization of feature-level supervision remain underexplored, largely due to limited annotated data. In this study, we introduce the STN dataset of 600 thyroid nodules with paired transverse and longitudinal ultrasound images, bounding box annotations, and complete labels for all five TI-RADS feature categories. Following the clinical decision process, we investigate how structured feature information can guide representation learning during training while requiring only images at inference. We demonstrate that text embeddings derived from standardized feature descriptions form a stable surrogate representation for TI-RADS risk levels. Based on this observation, we propose CMCNet, which aligns image embeddings to fixed textual embeddings via a Center-Margin Contrastive Loss that simultaneously promotes intra-class compactness and inter-class separation. Experimental results show that this embedding alignment strategy is more data-efficient and robust than direct multitask learning, and consistently outperforms InfoNCE, center loss, a strong multitask baseline, and a VQA-style multimodal model, particularly in imbalanced settings. The dataset is freely available at doi: 10.5281/zenodo.19125693 and the source code is available at: https://www.healthinformaticslab.org/supp/.

## 1 Introduction

Thyroid nodules are highly prevalent in the general population, and ultrasound is the primary imaging modality for their initial evaluation and follow-up (Bernet and Chindris 2021, Das *et al.* 2024). To standardize reporting and reduce inter-observer variability, the American College of Radiology (ACR) proposed the Thyroid Imaging

Reporting and Data System (TI-RADS), which assigns point-based scores to five key ultrasound feature categories (composition, echogenicity, shape, margin, and echogenic foci) and aggregates these scores into discrete risk levels (TR1–TR5) that guide biopsy and surveillance decisions (Tessler *et al.* 2017). In routine clinical practice, radiologists assess each attribute on both transverse and longitudinal views, compute the cumulative TI-RADS score, and determine a management strategy. Although clinically effective, this workflow is subjective, cognitively demanding, and time-consuming, particularly in high-throughput environments.

With the rapid development of deep learning (DL) in medical imaging, numerous studies have explored convolutional neural networks (CNNs) and related architectures for thyroid ultrasound analysis. Most existing models are formulated as binary classification or as a proxy task for biopsy recommendation, and report high area under the curve (AUC) values on curated datasets. For example, Peng et al. developed a multi-center DL system for differentiating benign from malignant nodules and demonstrated performance comparable to experienced radiologists (Peng *et al.* 2021). Zhao et al. proposed a Feature Fusion ResNet for the benign versus malignant classification (Zhao *et al.* 2022). Systematic comparisons of popular CNN backbones, including ResNet, DenseNet, and EfficientNet, have further established strong performance on biopsy-confirmed cohorts (Nasr *et al.*, n.d.). Machine-learning systems based on handcrafted ultrasound features or radiomics have also been investigated (Cao *et al.* 2021, Lu *et al.* 2022, Du *et al.* 2024, Guerrisi *et al.* 2024). Despite these advances, the majority of existing approaches optimize binary decision-making rather than modeling the full TI-RADS risk spectrum.

Relatively few studies directly address TI-RADS–consistent multi-class prediction or explicit modeling of the five underlying feature categories. Wu et al. integrated ACR TI-RADS descriptors with CNN features to improve discrimination between high- and low-risk nodules (Wu *et al.* 2021). However, their formulation focused on a reduced subset of TI-RADS levels rather than full five-class prediction. From a methodological perspective, the TI-RADS scoring system defines a structured decision rule: the five feature categories provide intermediate supervisory signals whose linear aggregation determines the final risk level. Such hierarchical supervision potentially offers richer constraints than binary labels alone. Nevertheless, to our knowledge, no publicly available thyroid ultrasound dataset provides complete annotations for all five TI-RADS feature categories, which may partially explain the limited progress in TI-RADS–consistent multi-class modeling. To address this gap, we construct a new dataset, STN (Single Thyroid Nodule–TI-RADS 1–5), comprising 600 nodules with paired transverse and longitudinal ultrasound images, bounding boxes, complete annotations of all five TI-RADS feature categories, and corresponding risk levels. This dataset enables systematic investigation of TI-RADS–consistent five-class modeling under structured supervision.

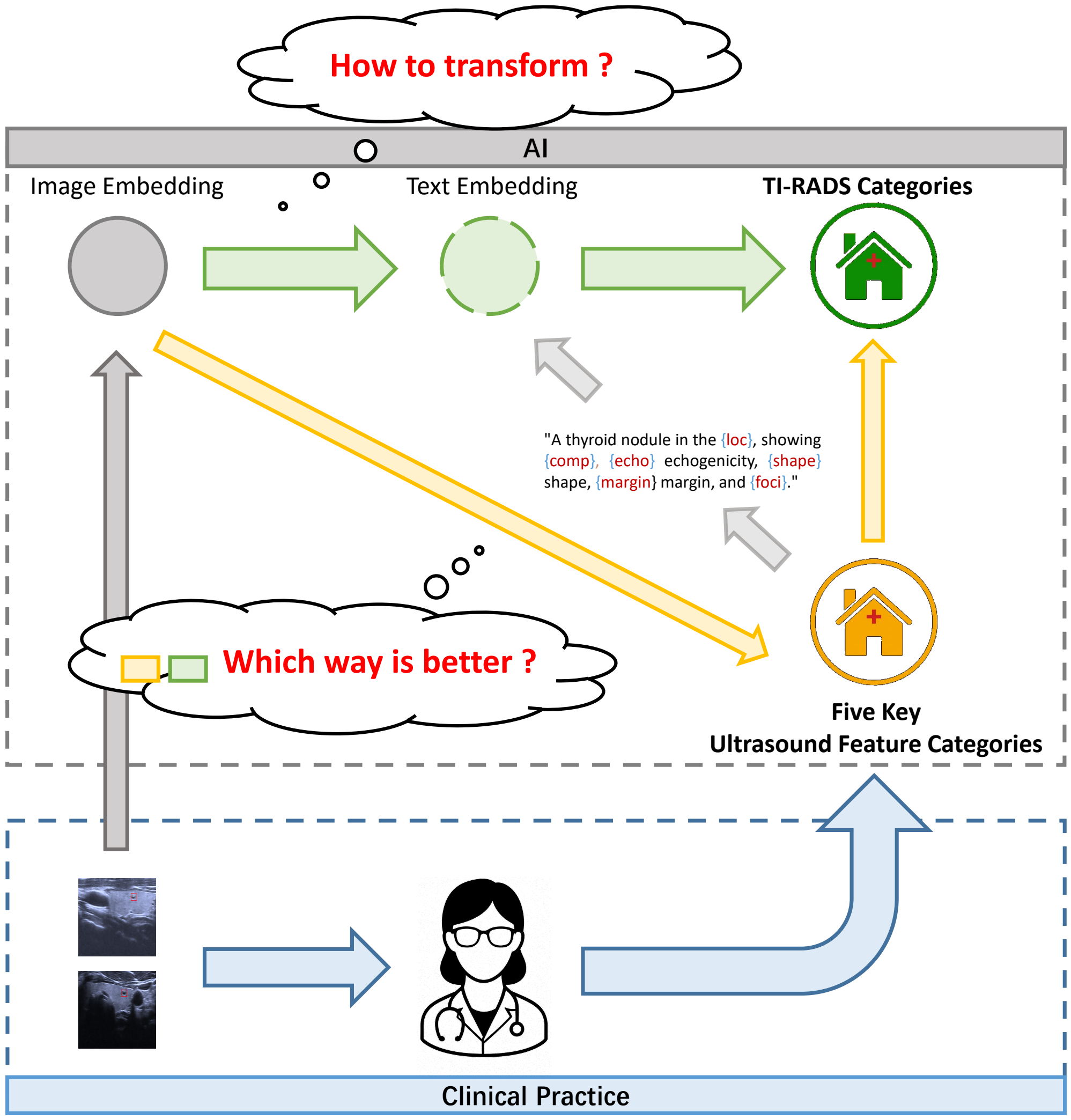

Figure 1. The basic idea of this paper.

Inspired by the clinical workflow, we aim to employ feature-level supervision during training while requiring only image inputs at inference. Radiologists evaluate the five TI-RADS attributes and compute the final category through linear aggregation. We hypothesize that these structured annotations encode a semantically stable representation of TI-RADS risk levels that can regularize image embedding learning. To test this, we convert the five feature categories into standardized textual descriptions and encode them using the frozen language model ModernBERT (Warner *et al.* 2025). The resulting text embeddings alone achieve near-perfect TI-RADS classification accuracy, and this indicates that structured textual representations form a compact surrogate space for risk discrimination. However, because inference in practice relies solely on images and feature annotations are scarce, our objective is to transfer this structured semantic information into image representations during training (see Figure 1).

We investigate two strategies. The first treats the five feature categories as auxiliary labels in a multi-head multitask framework. The second aligns image embeddings with fixed text embeddings derived from feature descriptions. While contrastive objectives such as InfoNCE (Oord, Li, and Vinyals 2019a, Chen *et al.* 2020, Liu S *et al.* 2022, Zhang R *et al.* 2026) used in the frameworks like GraphMVP (Liu S *et al.* 2022) and PepHarmony (PepHarmony 2026) encourage cross-modal discrimination, they do not explicitly enforce tight alignment between corresponding image–text pairs. To address this, we propose a unified objective combining margin loss (Chopra, Hadsell, and LeCun 2005a) and center loss (Wen *et al.* 2016), termed Center-Margin Contrastive Loss (CMCLoss), which promotes intra-class compactness, inter-class separation, and explicit attraction to textual anchors. Given the smaller inter-text distances relative to image embeddings, an additional L1 regularization term is introduced to stabilize gradient dynamics during alignment. The resulting model CMCNet is proposed to leverage textual anchors during training but performs image-only inference.

For comparison, we construct a strong multitask baseline, FLaMM-Net (Focal Loss–based Multi-view Multitask Network). Although TI-RADS categories are balanced in STN, the five feature categories are highly imbalanced. Therefore, focal loss is adopted for each auxiliary task head. Experimental results show that CMCLoss yields more stable and discriminative embedding alignment than InfoNCE and center loss, and that CMCNet consistently outperforms FLaMM-Net. Moreover, CMCNet surpasses a representative VQA-style multimodal model, LLaVA-Med, particularly under imbalanced conditions. These results suggest that for small-scale medical datasets governed by structured decision rules, alignment to fixed semantic anchors provides a more effective inductive bias than conventional multitask optimization or generative multimodal paradigms.

# 2 Related Work

Most existing studies formulate TI-RADS prediction as a binary classification task. To our knowledge, this work presents the first publicly reported thyroid ultrasound dataset annotated with point-based scores for all five TI-RADS feature categories together with the corresponding overall TI-RADS levels. In addition, we are the first to explicitly incorporate multimodal alignment between ultrasound images and structured textual representations for TI-RADS classification. In this section, we briefly review the image encoders, text encoder, and loss functions relevant to our framework.

## 2.1 Image Encoder Models

**BiomedCLIP** is a multimodal foundation model developed for biomedical vision-language tasks (Zhang S *et al.* 2025). It extends the CLIP framework by aligning medical images (e.g., X-ray, CT, MRI, microscopy, histology, pathology) with textual descriptions such as reports and figure captions through contrastive learning in a shared embedding space. Its architecture follows the standard CLIP dual-encoder design but is optimized for biomedical domains. However, ultrasound images are not included in its pretraining corpus.

**USFM** (Universal Ultrasound Foundation Model) is the first foundation model pre-trained specifically on ultrasound data (Jiao *et al.* 2024). It aims to support multiple organs and downstream tasks, including segmentation, classification, and image enhancement, with high label efficiency. USFM adopts a self-supervised masked image modeling strategy tailored to ultrasound images, using a spatial-frequency dual-masked modeling mechanism.

SAMUS is a domain-adapted version of the general-purpose Segment Anything Model (SAM) for ultrasound image segmentation (Lin X *et al.* 2024). By incorporating ultrasound-specific priors and adaptation strategies, SAMUS improves segmentation robustness on low-contrast and speckle-noise-dominated images.

To our knowledge, this work provides the first systematic evaluation of these recent foundation and transformer-based models for TI-RADS multi-class classification on thyroid ultrasound images. In addition to foundation models, conventional CNN-based backbones such as ResNet, DenseNet, and EfficientNet remain widely used in medical image analysis (Nasr *et al.*, n.d.). ResNet employs residual connections to enable deeper architectures, DenseNet promotes feature reuse through dense connectivity, and EfficientNet balances network depth, width, and input resolution through compound scaling. In thyroid malignancy prediction tasks on biopsy-confirmed cohorts, these architectures often achieve strong diagnostic performance (Nasr *et al.*, n.d.). However, direct comparisons between such CNN backbones and recent ultrasound-specific or multimodal transformer-based encoders (e.g., BiomedCLIP, USFM, SAMUS) for TI-RADS multi-class prediction have not been systematically reported.

## 2.2 Text Encoder Model

ModernBERT is a bidirectional, encoder-only Transformer model within the BERT family, and it is designed for efficient general-purpose language understanding (Warner *et al.* 2025). Compared with classical BERT architectures, it incorporates rotary positional embeddings, gated linear units, and an alternating local-global attention mechanism to support longer context windows while maintaining computational efficiency. Pre-trained on large-scale English text and code corpora, ModernBERT provides strong sentence-level and document-level representations for tasks such as retrieval and classification.

In this work, ModernBERT is used as a frozen text encoder to generate embeddings for standardized TI-RADS feature descriptions. Empirically, text embeddings alone achieve approximately 98% accuracy on both validation and test sets, indicating that structured textual representations provide a highly separable surrogate space for TI-RADS risk levels. Given this strong performance, additional text encoders were not evaluated.

## 2.3 Loss Functions

**InfoNCE loss** was introduced to optimize representation learning by maximizing mutual information between paired variables (Oord, Li, and Vinyals 2019b). It has since become a standard objective in contrastive learning frameworks such as SimCLR (Chen et al. 2020) and MoCo (He et al. 2020). The loss is defined as:

$$\mathcal{L}_{info} = -\log\left(\frac{e^{\frac{sim(x_i\ ,x_i^+)}{\mathcal{T}}}}{\sum_{i=1}^{M} e^{\frac{sim(x_i^+,x_j^-)}{\mathcal{T}}}}\right) \quad (1)$$

where $x_i$ denotes the representation of the $i^{\text{th}}$ sample, and $x_i^+$ is its positive counterpart (e.g., an augmented view), $x_j^-$ are negative samples. $\mathcal{T}$ is a temperature parameter that controls the sharpness of the similarity. $M$ is the number of negative samples.

**Center loss** was proposed by Wen et al. to enhance feature discriminability by reducing intra-class variance (Wen et al. 2016). It has been widely adopted in face recognition (Wen et al. 2019). The formulation is:

$$\mathcal{L}_{Cen} = \frac{1}{2N}\sum_{i=1}^{N} \boldsymbol{Dist}(x_i - c_{y_i}) \quad (2)$$

where $x_i$ is the features of the $i^{\text{th}}$ sample, and $c_{y_i}$ denotes the feature center of class $y_i$, and $N$ is the number of samples. The distance metric $Dist()$ is usually chosen as the L2 distance.

**Contrastive loss** was originally introduced for Siamese networks (Chopra, Hadsell, and LeCun 2005b). It pulls samples from the same class closer together while pushing samples from different classes apart:

$$\mathcal{L}_{contrast} = \frac{1}{2N}\sum_{i=1}^{N} (1 - y_i) \times \max\big(0, margin - \boldsymbol{Dist}(x_{1i}, x_{2i})\big) + y_i \times \boldsymbol{Dist}(x_{1i}, x_{2i}) \quad (3)$$

where $x_{1i}$, $x_{2i}$ are the $i^{\text{th}}$ pair of samples, $y_i$ indicates whether the pair belongs to the same class (1 for yes and 0 for no), The distance metric $\boldsymbol{Dist}()$ is usually chosen as the squared L2 distance. The $margin$ is a constant to control the minimum distance between samples from different classes.

**Focal loss** was introduced to address class imbalance by down-weighting easy samples and focusing training on hard examples (Lin TY *et al.* 2018). For multi-class classification:

$$\mathcal{L}_{focal} = -(1 - p_i)^{\gamma} \log(p_i) \quad (4)$$

where $p_i$ is the predicted probability of the ground-truth class and $\gamma$ controls the strength of down-weighting.

**Consistency loss** encourages prediction stability under input perturbations and is commonly used in robust or semi-supervised learning (Tarvainen and Valpola 2017, Miyato *et al.* 2018a, 2018b, Wei *et al.* 2020, Xie *et al.* 2020). Following JoCoR (Wei et al. 2020), we adopt a symmetric Kullback–Leibler divergence:

$$\mathcal{L}_{consist} = KL\left(P_{v_1} \parallel P_{v_2}\right) + KL\left(P_{v_2} \parallel P_{v_1}\right) \quad (5)$$

where $P_{v_1}$ and $P_{v_2}$ denote predictions from two perturbed views of the same input.

# 3 Materials and Methods

## 3.1 Dataset Descriptions

The STN (Single Thyroid Nodule with TI-RADS 1–5) dataset was retrospectively collected from the Department of Ultrasound at the China-Japan Union Hospital of Jilin University, covering the period from Jan, 2023 to Dec, 2025. All examinations were performed using high-frequency linear-array transducers on clinically approved ultrasound systems (Siemens ACUSON S3000, frequency range: 9-18 MHz). Only cases with complete transverse and longitudinal views of a single dominant thyroid nodule were included. For patients with multiple nodules, the clinically most suspicious nodule was selected according to routine diagnostic practice.

Each case includes paired transverse and longitudinal ultrasound images, manually annotated bounding boxes delineating the nodule region, and complete labels for all five TI-RADS feature categories: composition, echogenicity, shape, margin, and echogenic foci. Based on the ACR TI-RADS guideline (Tessler *et al.* 2017), each feature category was assigned a point-based score, and the total score was obtained by linear summation to determine the final TI-RADS level (TR1–TR5), as illustrated in Figure 2 (a).

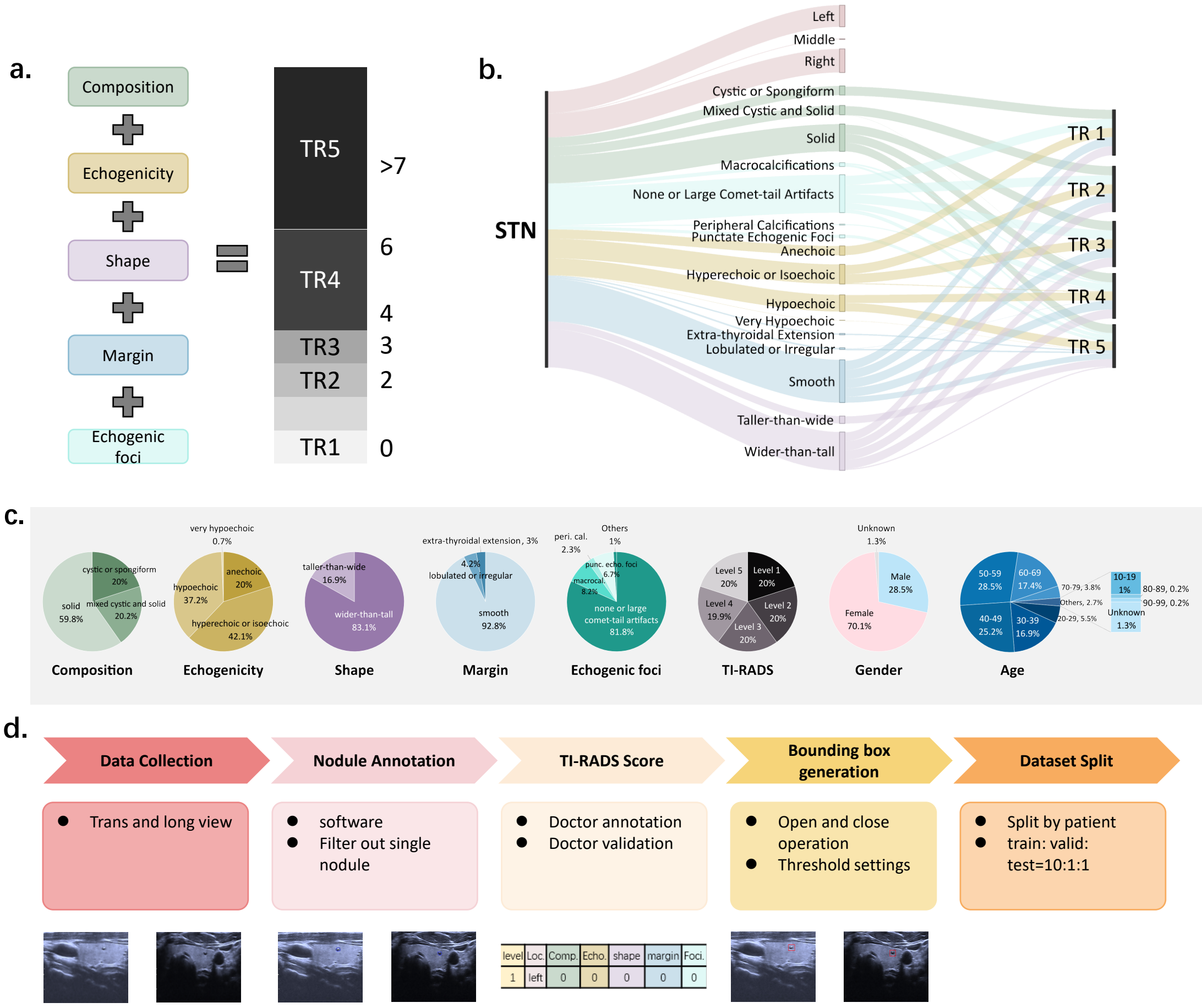

**Figure 2.** Overview of the STN dataset. (a) The five key TI-RADS feature categories and their linear aggregation into the final TI-RADS score. (b) Data processing pipeline of the STN dataset. (c) Distribution of composition, echogenicity, shape, margin, echogenic foci, TI-RADS categories, age, and gender. (d) Workflow for constructing the STN dataset.

All TI-RADS feature annotations and final risk levels were independently assessed by at least two experienced ultrasound physicians with more than 10 years of thyroid ultrasound experience. In cases of disagreement, a consensus reading was conducted. The detailed annotation workflow is summarized in Figure 2 (d). Briefly, the process included: (1) case screening and eligibility verification, (2) image quality control and view confirmation, (3) feature-level scoring according to ACR TI-RADS criteria, (4) bounding box annotation of the nodule region, and (5) dataset curation and anonymization.

Figure 2 (b) illustrates the data organization pipeline, including raw image acquisition,

nodule localization, feature-level annotation, and structured dataset assembly. The distribution of TI-RADS categories, individual feature attributes, and patient demographics (age and gender) is presented in Figure 2 (c). Notably, while the overall TI-RADS categories are approximately balanced, substantial class imbalance exists within specific feature categories, particularly in echogenic foci or shape, reflecting real-world clinical prevalence.

This study was approved by the Institutional Review Board of China-Japan Union Hospital of Jilin University (Approval No. 2026030504). Owing to the retrospective design and the use of fully anonymized patient data, the requirement for written informed consent was waived.

## 3.2 Overall Architectures

For the proposed multimodal alignment model CMCNet, we formalize the training procedure as follows (Figure 3). Figure 3 (a) presents the overall architecture of CMCNet. Longitudinal (L-view) and transverse (T-view) ultrasound images are encoded by a shared image encoder, pooled, and fused into a patient-level image representation. Structured TI-RADS feature scores are converted into standardized textual descriptions and encoded by a frozen text encoder. Both modalities are projected into a shared embedding space and aligned using the proposed Center-Margin Contrastive Loss (CMCLoss). Figure 3 (b) illustrates the FLaMM-Net baseline. Similar two-view encoding and fusion are applied, followed by multi-head classifiers optimized with focal loss. The sum of focal losses across views is defined as the auxiliary loss. Consistency loss is included only for ablation analysis and is not part of the final model. Figure 3 (c) compares two multi-head formulations. Multi-head version 1 predicts all tasks from a shared latent representation. Multi-head version 2 first learns feature-specific latent representations and then concatenates them to predict the final TI-RADS category, mimicking the clinical scoring workflow. Figure 3 (d) shows representative prediction results under different configurations. The BiomedCLIP setting corresponds to Multi-head v1 with L2-based CMCLoss, while the USFM setting corresponds to Multi-head v2 with L1-based CMCLoss.

The technical details are defined in the following sections.

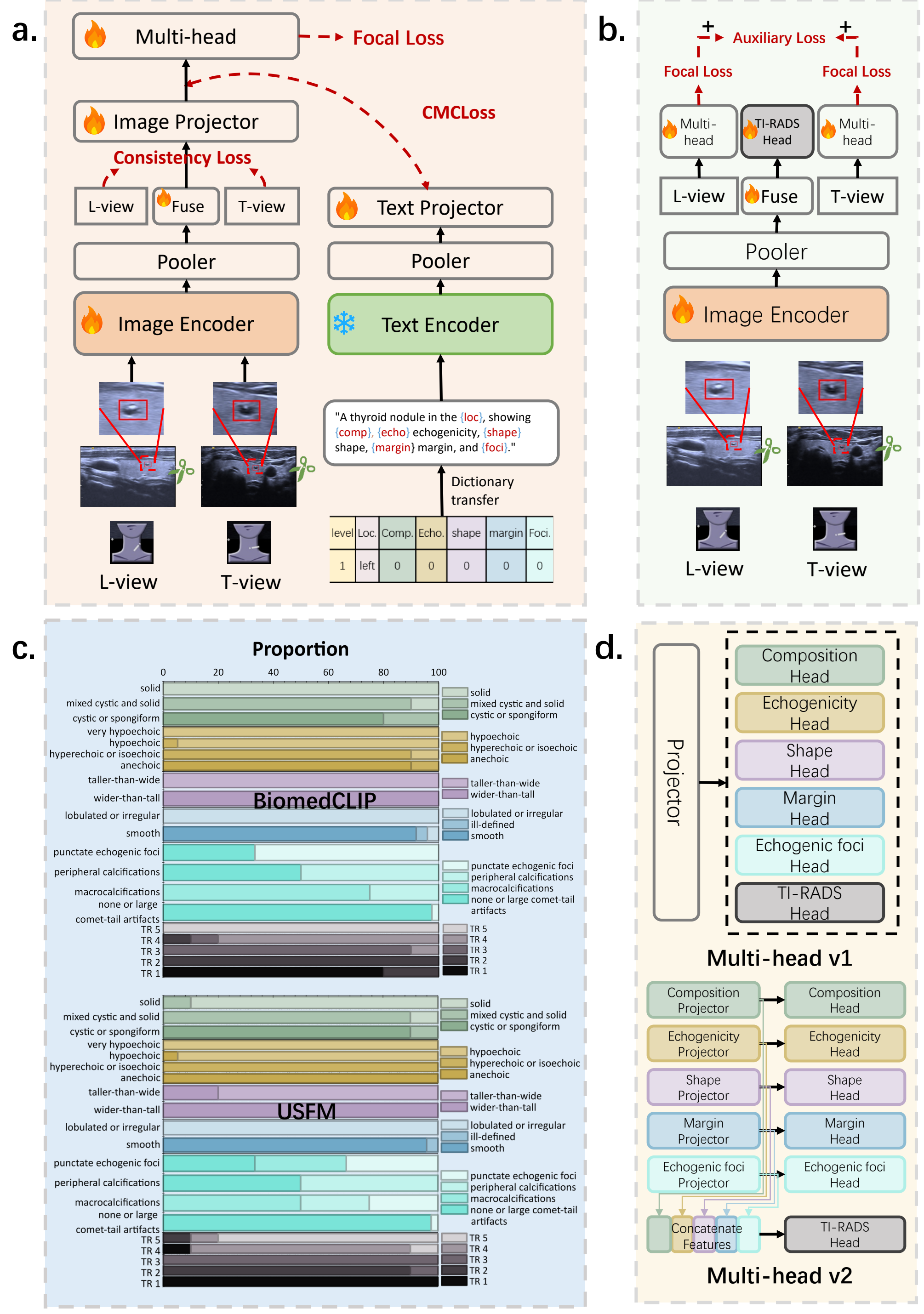

Figure 3. Model architectures and design variants. (a) CMCNet framework with multimodal alignment via CMCLoss. (b) FLaMM-Net baseline with focal-loss multitask learning. (c) Representative results: for BiomedCLIP, using multi-head v1 and L2-based CMC Loss, for USFM, using multi-head v2 and L1-based CMC Loss. (d) Multi-head v1 (shared latent) vs. Multi-head v2 (hierarchical feature-driven).

### 3.3 Embedding Definitions and Model Variants

The training procedure of the CMCNet model is defined as follows (see Figure 3 (a)).

**Definitions of image embeddings.** Given a patient $i$, we have two ultrasound images: a transverse view $I_i^t$ and a longitudinal view $I_i^l$. For an image encoder $\boldsymbol{IE}$, we obtain patch-level image embeddings $\boldsymbol{IE}(I_i^t) \in R^{n \times h_I}$ and $\boldsymbol{IE}(I_i^l) \in R^{n \times h_I}$, where $n$ is the number of patches, $h_I$ is the hidden dimension of the image embeddings. After a pooling operation, we obtain the view-level image embeddings $F_{I_i}^t = pooler\left(\boldsymbol{IE}(I_i^t)\right), F_{I_i}^t \in R^{h_I}$, so it does for $F_{I_i}^l$. Then we define a fusion function $\boldsymbol{Fuse_I}$ to merge the two view-level embeddings, yielding the final image embedding $F_{I_i} = \boldsymbol{Fuse_I}(F_{I_i}^t, F_{I_i}^l)$.

**Definitions of text embeddings.** Given any patient $i$, we have point-based TI-RADS scores for the five key ultrasound feature categories: composition, echogenicity, shape, margin, and echogenic foci. We denote them as $S_{comp}, S_{\text{echo}}, S_{shape}, S_{margin}$, and $S_{foci}$ respectively, and we collect the location of the nodule, denoted by $S_{loc}$. By projecting the score into descriptions by a dictionary $Dict$ (the details of $Dict$ can be found in the Appendix), we transform the structured data into text by a Python functional string template:

*f"A thyroid nodule in the {$Dict(S_{loc})$}, showing {$Dict(S_{comp})$}, {$Dict(S_{echo})$} echogenicity, {$Dict(S_{shape})$} shape, {$Dict(S_{margin})$} margin, and {$Dict(S_{foci})$}."*

By substituting the variables in the template with their corresponding descriptions, we get the image caption $T_i$ for each patient $i$. Then we get the text embedding $\boldsymbol{TE}(T_i) \in R^{m \times h_T}$ using a language model $\boldsymbol{TE}$, where $m$ is the number of tokens, $h_T$ is the hidden dimensions of the text embeddings. After a pooling operation, we get the final text embedding $F_{T_i} = pooler\left(\boldsymbol{TE}(T_i)\right), F_{T_i} \in R^{h_T}$.

**Definitions of the shared embedding space.** Given a patient $i$, we have the final image embedding $F_{I_i} \in R^{h_I}$ and the final text embedding $F_{T_i} \in R^{h_T}$. To map them into the same shared embedding space, we define an image projector $\boldsymbol{P_I}$ and a text projector $\boldsymbol{P_T}$, inspired by canonical correlation analysis (CCA) (Hardoon, Szedmak, and Shawe-Taylor 2004). They can transform both image and text embeddings into a shared embedding space with the same dimension, namely $G_{I_i} = \boldsymbol{P_I}(F_{I_i})$, $G_{I_i} \in R^{h_S}$ and $G_{T_i} = \boldsymbol{P_T}(F_{T_i})$, $G_{T_i} \in R^{h_S}$.

**Definitions of the Multi-head classifiers version 1 (Multi-head v1).** This model follows a standard multitask learning setup. Given a patient $i$, we predict TI-RADS and the five key features categories together, namely $\widehat{S_*} = \boldsymbol{Head}_*(\boldsymbol{P_h}(F_{I_i}))$, where $\widehat{S_*}$ is the predicted score and $\boldsymbol{P_h}$ is a shared projector, $\boldsymbol{Head}_*$ is a classifier. * can be one of the five TI-RADS categories, i.e., composition, echogenicity, shape, margin, echogenic foci.

**Definitions of the Multi-head classifiers version 2 (Multi-head v2).** This version is motivated by the workflow commonly adopted in clinical diagnostic practice. Given a patient $i$, we first project and predict the each of the five key features categories in a shared latent space, $\widehat{F_{I_{*i}}} = \boldsymbol{P}_*(F_{I_i}), \widehat{S_*} = \boldsymbol{Head}_*(\widehat{F_{I_{*i}}})$, where $\widehat{F_{I_{i*}}} \in R^{h_p}$, $h_p$ is the hidden dimension of each task-specific head, and * indexes one of the five TI-RADS categories, i.e., composition, echogenicity, shape, margin, or echogenic foci. Then concatenate these head-specific features $\widehat{F_{I_{i_{tirads}}}} = [\widehat{F_{I_{comp_i}}} \parallel \widehat{F_{I_{echo_i}}} \parallel \widehat{F_{I_{shape_i}}} \parallel \widehat{F_{I_{margin_i}}} \parallel \widehat{F_{I_{foci_i}}}]$, $\widehat{F_{I_{i_{tirads}}}} \in R^{5h_p}$ And then we build a TI-RADS head classifier to predict TI-RADS category $\widehat{S_{tirads}} = \boldsymbol{Head}_*(\widehat{F_{I_{i_{tirads}}}})$.

## 3.4 Loss Functions

After introducing the core model components, we describe the loss functions used for training. We observe that the InfoNCE objective tends to reduce the loss by continuously increasing inter-sample distances among negative pairs (Schrodi *et al.* 2025, Liang *et al.*, n.d.). This behavior is suboptimal for our setting, where the primary goal is to explicitly pull each image embedding toward its corresponding textual representation.

In our framework, the five TI-RADS feature categories (composition, echogenicity, shape, margin, and echogenic foci) are converted into text using a fixed caption template. As a result, samples sharing identical feature scores naturally correspond to a common text-embedding center. This observation motivates the introduction of a center-based alignment term. At the same time, we empirically find that distances between distinct text embeddings are substantially smaller than those among image embeddings (Supplementary Materials), which necessitates an explicit mechanism to maintain inter-class separation. We therefore combine a center loss with a margin-based contrastive term, resulting in the proposed **Center-Margin Contrastive Loss (CMCLoss)**.

**Definition of Center-Margin Contrastive Loss.** Given a patient $i$, we get the shared embeddings of images and text, $G_{I_i}, G_{T_i} \in R^{hs}$. Since we convert the five key ultrasound feature categories to text using a fixed caption template, samples with the same scores for the five TI-RADS categories share the same text embedding center $C_{y_i}$. The definition of CMCLoss as follows:

$$\mathcal{L}_{CMC} = w_o \times max\left(0, margin - \boldsymbol{Dist}\left(G_{I_i}, C_{y_j}\right)\right) + w_1 \times \boldsymbol{Dist}\left(G_{I_i}, C_{y_j}\right) \quad (6)$$

$$C_{y_j} = \frac{1}{m}\sum_{i=1}^{m} G_{T_i} \quad (7)$$

where $G_{I_i}$ and $C_{y_j}$ are the image embedding and the text embedding center of the $i^{\text{th}}$ patient, respectively. The constant $margin$ is used to control the minimum distance between samples of different classes. This study calculates the distance metric $\boldsymbol{Dist}()$ using the L2 distance and the L1 distance. $w_o$ and $w_1$ are weighting coefficients that balance the contributions of the inter-class and intra-class terms.

**Definition of focal loss.** Although the TI-RADS categories are balanced in the dataset, the five key ultrasound feature categories are extremely imbalanced. To solve these imbalanced classification problems, we use focal loss to constrain the five feature-classification heads (the same as Eq. 4):

$$\mathcal{L}_{focal} = -\left(1 - \widehat{S}_*\right)^{\gamma} \log\left(\widehat{S}_*\right) \tag{8}$$

In FLaMM-Net, the loss is computed independently for each view:

$$\mathcal{L}_{aux} = -\left(1 - \widehat{S}_*^t\right)^{\gamma} \log\left(\widehat{S}_*^t\right) - \left(1 - \widehat{S}_*^l\right)^{\gamma} \log\left(\widehat{S}_*^l\right) \tag{9}$$

where $\widehat{S}_*^t$ and $\widehat{S}_*^l$ denote the predicted scores of transverse and longitudinal views, respectively. For convenience, we refer to this term as the auxiliary loss.

Definition of consistency loss. We use consistency loss to encourage agreement between the multi-view features from the same patient. The loss function has the same form as in Eq. 5:

$$\mathcal{L}_{consist} = KL\left(\widehat{S}_*^t \parallel \widehat{S}_*^l\right) + KL\left(\widehat{S}_*^l \parallel \widehat{S}_*^t\right) \tag{10}$$

# 4 Experimental Results

Detailed experimental settings for each model, including implementation configurations and data-splitting strategies, are provided in the Supplementary Materials. Unless otherwise specified, all results reported in this section are evaluated on the test set of the STN dataset.

## 4.1 Text Modality Provides a Near-Upper-Bound Representation

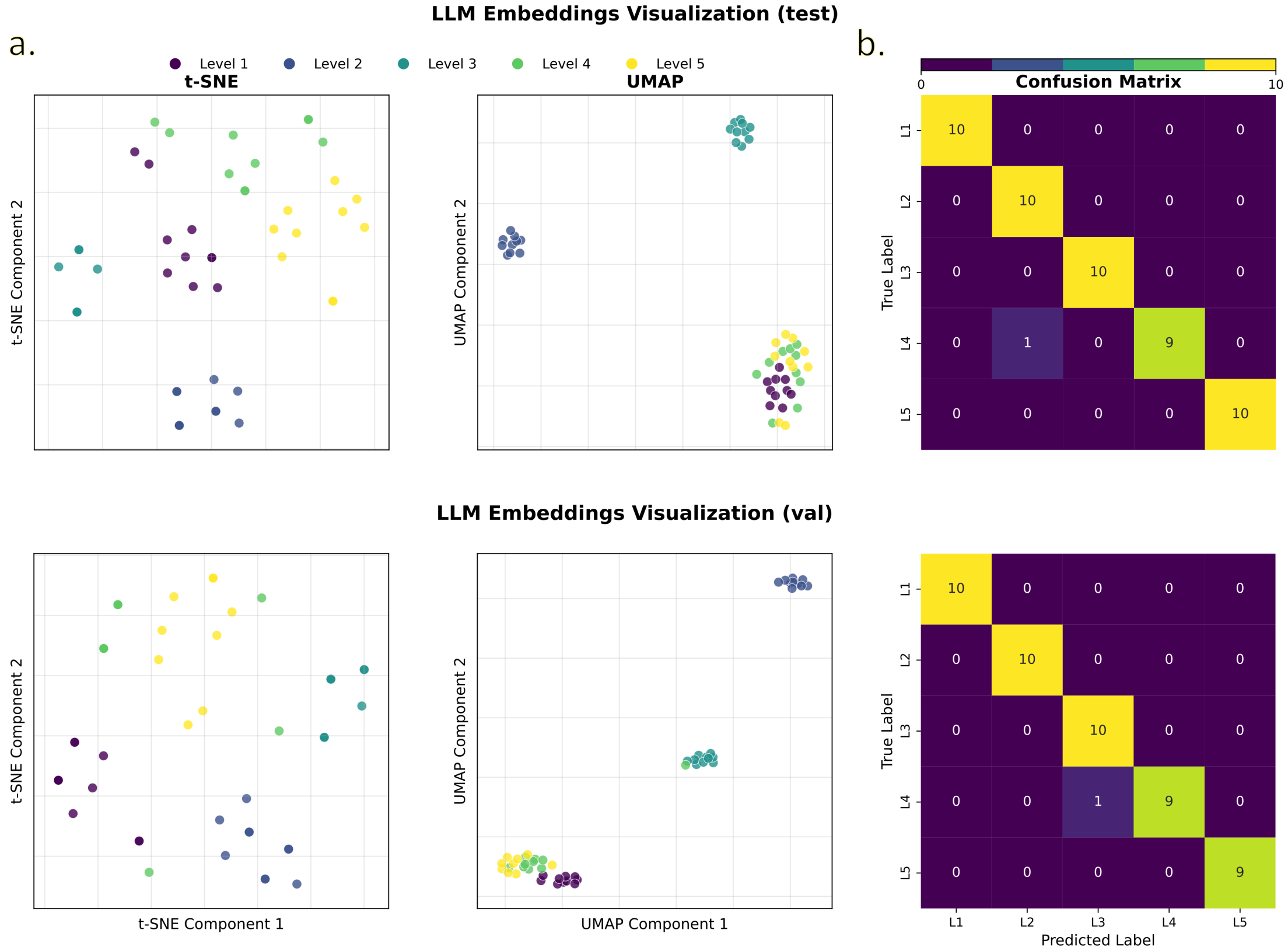

**Figure 4. Discriminative capacity of the text modality.** (a) Two-dimensional visualization of pooled ModernBERT text embeddings using t-SNE and UMAP, color-coded by TI-RADS category. (b) Confusion matrix of the text-only classifier on the test set.

To evaluate the discriminative power of structured textual representations, we encode template-generated TI-RADS captions using a frozen ModernBERT encoder (Warner et al. 2025) without fine-tuning. A single fully connected classification head is trained on the pooled text embeddings using the STN training set, with model selection based on validation performance. The model converges within five epochs.

As shown in Figure 4 (a), both t-SNE and UMAP projections reveal clear and well-separated clusters corresponding to the five TI-RADS risk levels. The embeddings exhibit minimal inter-class overlap, and the structured captions derived from feature-level scores demonstrate a compact and linearly separable representation space.

Quantitatively, the text-only classifier achieves 97.5% accuracy on the validation set and 98.0% accuracy on the test set for the TI-RADS risk-level prediction. The confusion matrix in Figure 4 (b) further confirms near-perfect diagonal dominance, with only marginal misclassifications between adjacent risk levels.

These findings demonstrate that once the five category-level scores are known, the corresponding textual representation provides an almost deterministic mapping to the overall TI-RADS risk levels. The embedding space defined by the pretrained ModernBERT representation model therefore functions as a near-upper-bound representation under perfect category feature annotation.

This behavior aligns with prior results in vision–language pretraining, where frozen language encoders often produce highly separable representations when textual inputs encode explicit semantic attributes (Radford *et al.* 2021, Li J *et al.* 2023). In our context, the near-ceiling performance supports the central hypothesis that structured feature supervision defines a semantically stable surrogate space. The subsequent objective is therefore to align image embeddings toward this structured textual anchor space while maintaining image-only inference capability.

## 4.2 Direct Optimization of the Target from Images

### 4.2.1 Image-Only Baselines and Ablation Design

This section evaluates how well TI-RADS prediction can be optimized directly from ultrasound images when feature-level annotations are used as auxiliary supervision. We construct FLaMM-Net as a unified image-only framework (Figure 3 (b)). The model takes either a single transverse view or paired transverse–longitudinal views, and optionally crops inputs using the annotated nodule bounding box.

To reflect the structured TI-RADS scoring rule, we implement two multi-task classifier designs (Figure 3 (d)): Multi-head v1, which predicts all tasks from a shared latent representation, and Multi-head v2, which first learns feature-specific representations and then concatenates them to predict TI-RADS. Since the five feature categories are highly imbalanced, we adopt focal loss. For multi-view training, the focal losses of the two views are summed and referred to as the auxiliary loss.

### 4.2.2 Which Image Encoder Performs Best?

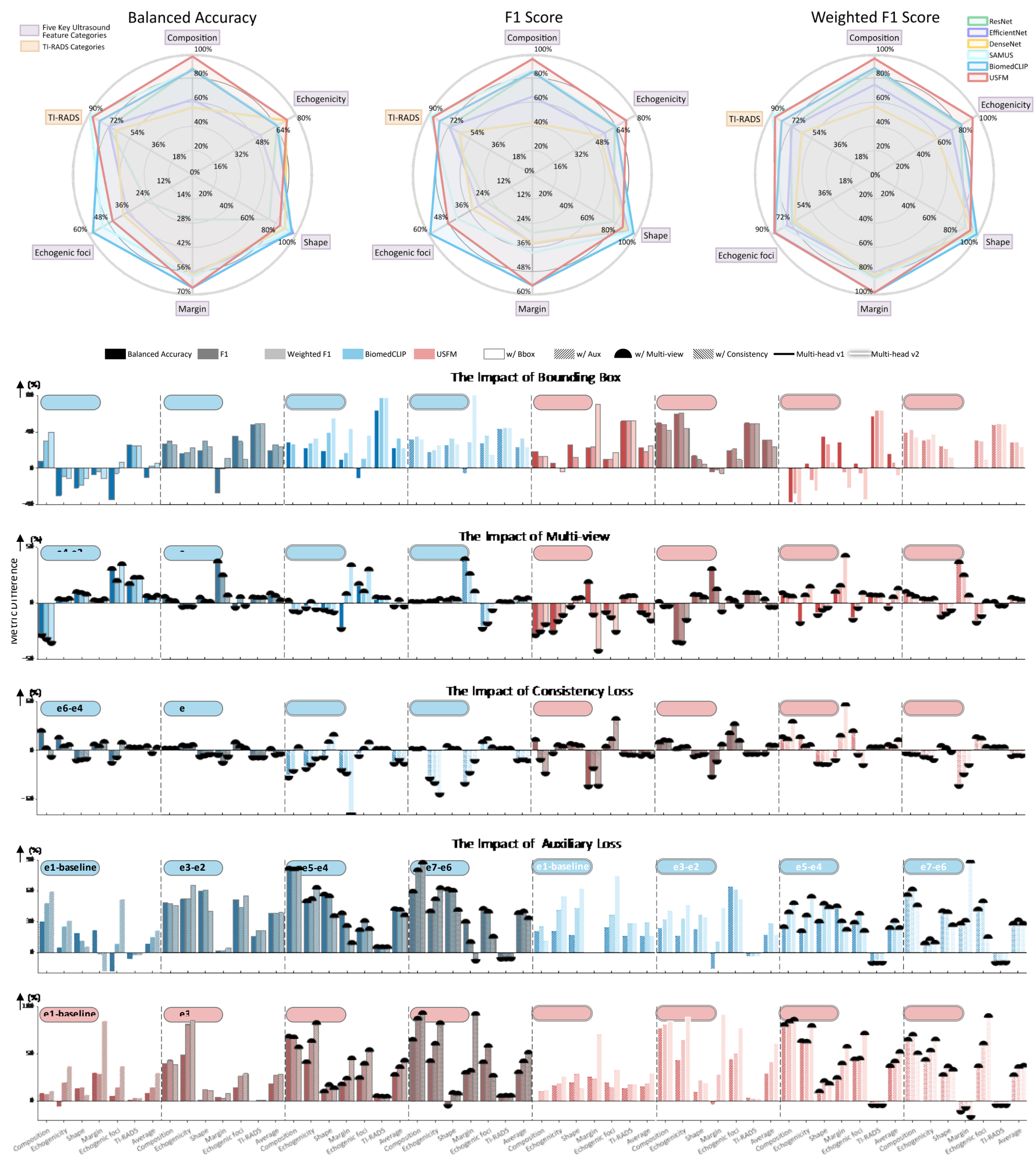

**Figure 5. Ablation analysis of FLaMM-Net for image-only TI-RADS learning.** (a) Comparison of six image encoders under the e5 configuration. (b–e) Module-wise ablations relative to a common baseline. The x-axis enumerates six prediction tasks and their average; the y-axis reports percentage change in evaluation metrics. Blue/red indicate BiomedCLIP/USFM, respectively. Solid/dashed lines denote Multi-head v1/v2.

We evaluate six encoders: BiomedCLIP, USFM, SAMUS, and three CNN backbones (DenseNet, EfficientNet, ResNet50). Figure 5 (a) shows that USFM and BiomedCLIP achieve the strongest overall performance across the six tasks. Specifically, USFM achieves 75.17% balanced accuracy, 73.80% F1, and 91.78% weighted F1, while BiomedCLIP reaches 74.81%, 73.15%, and 88.70%, respectively. BiomedCLIP yields the highest scores on shape and margin, whereas USFM performs best on composition and echogenicity. For echogenic foci, BiomedCLIP achieves the best balanced accuracy and F1, while USFM achieves the best weighted F1.

SAMUS achieves the best performance on the TI-RADS task but degrades substantially on several attribute heads, particularly margin and echogenic foci. Because TI-RADS categories are derived from the feature-level scoring rule, strong attribute prediction is a more reliable basis for TI-RADS modeling. Therefore, we select USFM and BiomedCLIP for subsequent ablations and analysis.

The results are consistent with evidence that foundation-model pretraining improves downstream medical imaging performance and label efficiency (Jiao *et al.* 2024, Liu Y *et al.* 2025).

### 4.2.3 Module-Wise Ablations of FLaMM-Net

**Table 1. Experimental configurations for the FLaMM-Net ablation study.** “Bbox” denotes bounding-box cropping of the nodule region; “Aux” denotes auxiliary supervision via summed view-wise focal losses. “✔” indicates that FLaMM-Net is trained with the corresponding module enabled.

| Experiments | Bbox | Multi-view | Consistency | Aux |
|---|---|---|---|---|
| **baseline** | | | | |
| **e1** | | | | ✔ |
| **e2** | ✔ | | | |
| **e3** | ✔ | | | ✔ |
| **e4** | ✔ | ✔ | | |
| **e5** | ✔ | ✔ | | ✔ |
| **e6** | ✔ | ✔ | ✔ | |
| **e7** | ✔ | ✔ | ✔ | ✔ |

To quantify the contribution of individual components, we perform a staged ablation study (Table 1). The baseline uses transverse view only, no bounding-box cropping, and optimizes TI-RADS only using cross-entropy. Modules are then introduced incrementally: auxiliary loss (e1), bounding-box cropping (e2), auxiliary + cropping (e3), multi-view (e4), multi-head (e5), consistency loss (e6), and consistency + auxiliary (e7).

Performance changes are summarized in Figure 5 (b–e) using balanced accuracy, F1, and weighted F1.

**Impact of bounding-box cropping.** Cropping the nodule region improves performance across most tasks for both USFM and BiomedCLIP (Figure 5 (b)), particularly when combined with auxiliary supervision. This indicates that restricting the input to the diagnostic region reduces background noise and stabilizes feature learning.

**Impact of multi-view input.** Multi-view input without auxiliary supervision yields mixed effects (Figure 5 (c)). Performance gains are most apparent for margin and TI-RADS, while simpler appearance attributes (e.g., composition, echogenicity) can degrade. Once auxiliary loss is introduced, multi-view becomes consistently beneficial across encoders and head variants. The data suggests that effective multi-view fusion depends on explicit feature-level supervision.

**Impact of consistency loss.** Consistency loss produces configuration-dependent changes and is not uniformly beneficial (Figure 5 (d)). In several settings (especially BiomedCLIP with Multi-head v2), the harmful effect suggests that enforcing strict agreement between views may suppress complementary information. This provides context for the multimodal alignment strategy in the next subsection, where intermediate representations are additionally constrained by text anchors.

**Impact of auxiliary loss.** Auxiliary loss provides the most stable improvements across settings (Figure 5 (e)). It consistently strengthens prediction of the five feature categories, and its benefit becomes more pronounced when the configuration is richer (bounding-box cropping and multi-view). TI-RADS performance occasionally shows minor trade-offs, which likely reflects the fact that directly optimizing feature heads does not guarantee optimal risk-level separation without additional structure in the representation space.

### 4.2.4 Summary of the Ablation Study

Across all configurations, three consistent patterns emerge. First, bounding-box cropping of the nodule region produces stable performance gains across most attribute heads and the TI-RADS task. Second, multi-view input is particularly beneficial for margin and TI-RADS prediction, but its effectiveness depends on auxiliary supervision. In the absence of auxiliary supervision, simpler appearance attributes such as composition and echogenicity may exhibit performance degradation. Third, auxiliary loss provides the most consistent and robust improvements across encoders and multi-head variants. In contrast, the consistency loss yields configuration-dependent effects and is not uniformly advantageous. These findings highlight the limitation of direct optimization from images alone and motivate the introduction of stronger intermediate constraints through multimodal alignment in the following section.

## 4.3 Evaluation of Intermediate Embedding Alignment

In the previous section, several modules (particularly Multi-head v2, auxiliary loss, and consistency loss) showed unstable or limited gains when optimization was performed directly from images. We hypothesized that this behavior arises from insufficient constraints on intermediate representations. CMCNet addresses this limitation by aligning image embeddings with structured textual embeddings for the latent space regularization. In this section, we re-evaluate auxiliary supervision, consistency regularization, and multi-head design under this alignment framework.

**Effect of auxiliary loss.** As shown in Figure 6 (a), auxiliary supervision produces more consistent improvements in CMCNet than in FLaMM-Net. For BiomedCLIP, nearly all configurations improve TI-RADS prediction and all improve feature-level tasks. Similar trends are also observed for USFM. These results indicate that once image embeddings are aligned to text-derived semantic centers, feature-level supervision more effectively stabilizes optimization and reduces ambiguity in the learned decision function.

**Effect of consistency loss.** Figure 6 (b) shows that consistency loss yields predominantly positive or neutral effects in CMCNet, in contrast to its unstable behavior in the image-only framework. Improvements are particularly obvious for margin and echogenic foci. When embeddings are constrained toward shared textual anchors, encouraging agreement between transverse and longitudinal views reduces intra-class variance and facilitates more reliable alignment.

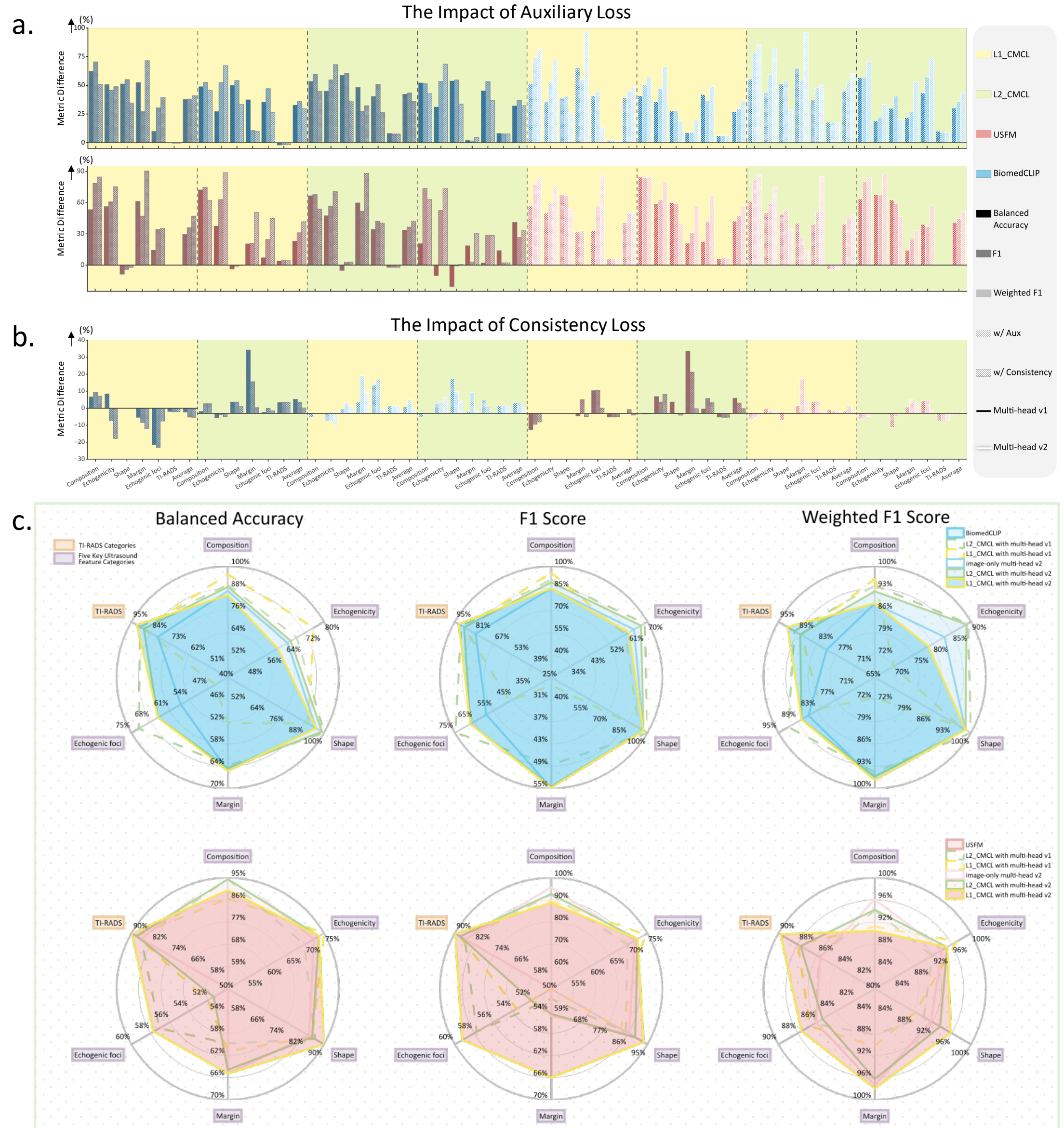

**Figure 6. The ablation studies of CMCNet.** the red and blue color represent USFM and BiomedCLIP respectively. The gradient bar colors indicate the three evaluation metrics, balanced accuracy, F1 score and weighted F1 score. (a) Auxiliary loss has a positive impact on the TI-RADS task in CMCNet. The y-axis shows the performance differences between models trained with and without auxiliary loss, and larger values indicate a stronger benefit of auxiliary loss. (b) Consistency loss has a positive impact on CMCNet. The y-axis shows the performance difference between models trained with and without consistency loss, and larger values indicate a stronger benefit of consistency loss. (c) The performance of multi-head v2 and multi-head v1 depends on the image encoders. For BiomedCLIP, multi-head v1 performs best, for USFM, multi-head v2 performs best. But CMCNet performs better than FLaMM-Net on the multi-head v2.

**Multi-head structure under alignment.** Figure 6 (c) demonstrates that the preferred multi-head structure depends on the encoder: BiomedCLIP favors Multi-head v1, whereas USFM benefits more from Multi-head v2. It's also interesting to see that under embedding alignment, Multi-head v2 no longer exhibits the degradation observed in FLaMM-Net and consistently outperforms its image-only counterpart. Overall, these findings support the hypothesis that aligning intermediate embeddings reduces representation uncertainty and enhances structural consistency with the TI-RADS scoring mechanism.

In the following section, we focus on the two strongest configurations for further analysis, i.e., BiomedCLIP with Multi-head v1 and USFM with Multi-head v2.

## 4.4 CMCLoss as an Effective Image–Text Alignment Objective

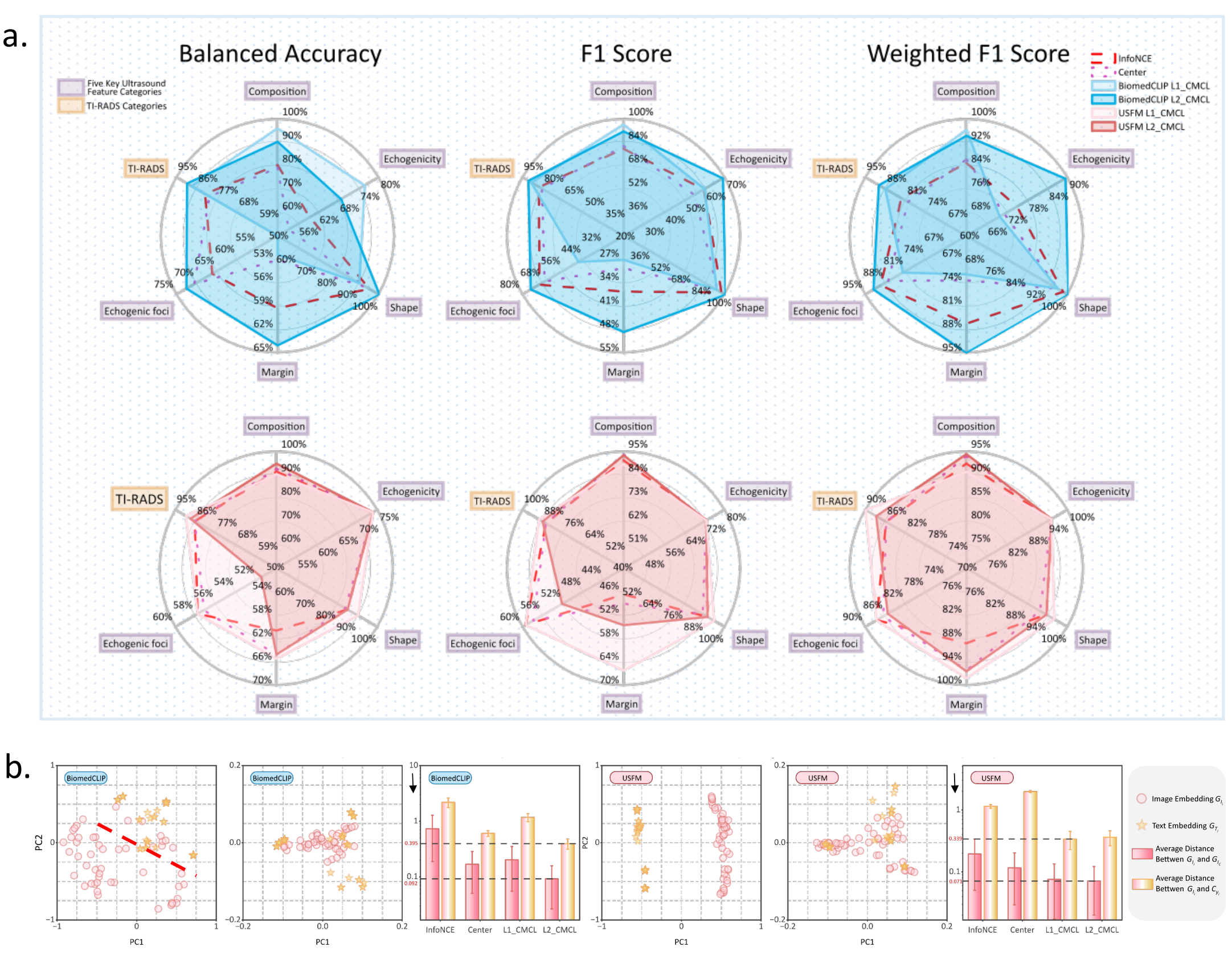

**Figure 7. Comparison of alignment objectives and embedding behavior.** (a) Performance comparison among InfoNCE, Center Loss, and CMCLoss (L1 and L2 variants). (b) PCA visualization of image and text embeddings in the shared space, together with quantitative distance analysis. Bar plots report (i) the average distance

between image embeddings and their image-class centers and (ii) the average distance between image embeddings and their corresponding text embedding centers. Smaller distances indicate stronger intra-class compactness and improved cross-modal alignment.

We introduce Center-Margin Contrastive Loss (CMCLoss) and evaluate its L1 and L2 variants against InfoNCE and Center Loss to assess their effectiveness for image-text alignment. Experiments are conducted under the two representative configurations identified in Section 4.3, namely BiomedCLIP with Multi-head v1 and USFM with Multi-head v2.

As shown in Figure 7 (a), CMCLoss achieves the strongest overall performance among the evaluated objectives. For BiomedCLIP with Multi-head v1, the L2-based variant attains the highest average performance and shows clear advantages on shape, margin, and echogenic foci. For USFM with Multi-head v2, the L1-based variant yields the best overall average and outperforms InfoNCE and Center Loss on five of six tasks. CMCLoss consistently improves task-level and averaged metrics across both configurations, indicating that jointly enforcing center aggregation and margin separation is more effective for structured cross-modal alignment than relying solely on contrastive separation or intra-class compactness.

To examine the geometric properties induced by different alignment objectives, we project image and text embeddings into a two-dimensional PCA space, as shown in Figure 7 (b). Models trained with InfoNCE exhibit a noticeable modality gap (particularly for USFM), where image and text clusters remain clearly separated. In contrast, CMCLoss produces tighter cross-modal aggregation and reduced intra-class dispersion. Quantitative analysis shows that configurations with smaller distances between image embeddings and their corresponding text centers consistently achieve stronger downstream performance. Among the evaluated objectives, L2_CMCL yields the lowest intra-class dispersion for both encoders, consistent with its explicit center-constraining formulation. By comparison, InfoNCE primarily emphasizes separation of negative pairs, which can increase inter-modality distance without guaranteeing alignment to shared semantic centers.

Overall, CMCLoss provides a more effective objective for transforming image embeddings into structured text-aligned representations. By jointly enforcing intra-class compactness and cross-modal alignment, it shapes an embedding space that is more consistent with the structured TI-RADS scoring framework and yields improved downstream classification performance.

## 4.5 Embedding Alignment vs. Direct Image-Only Optimization

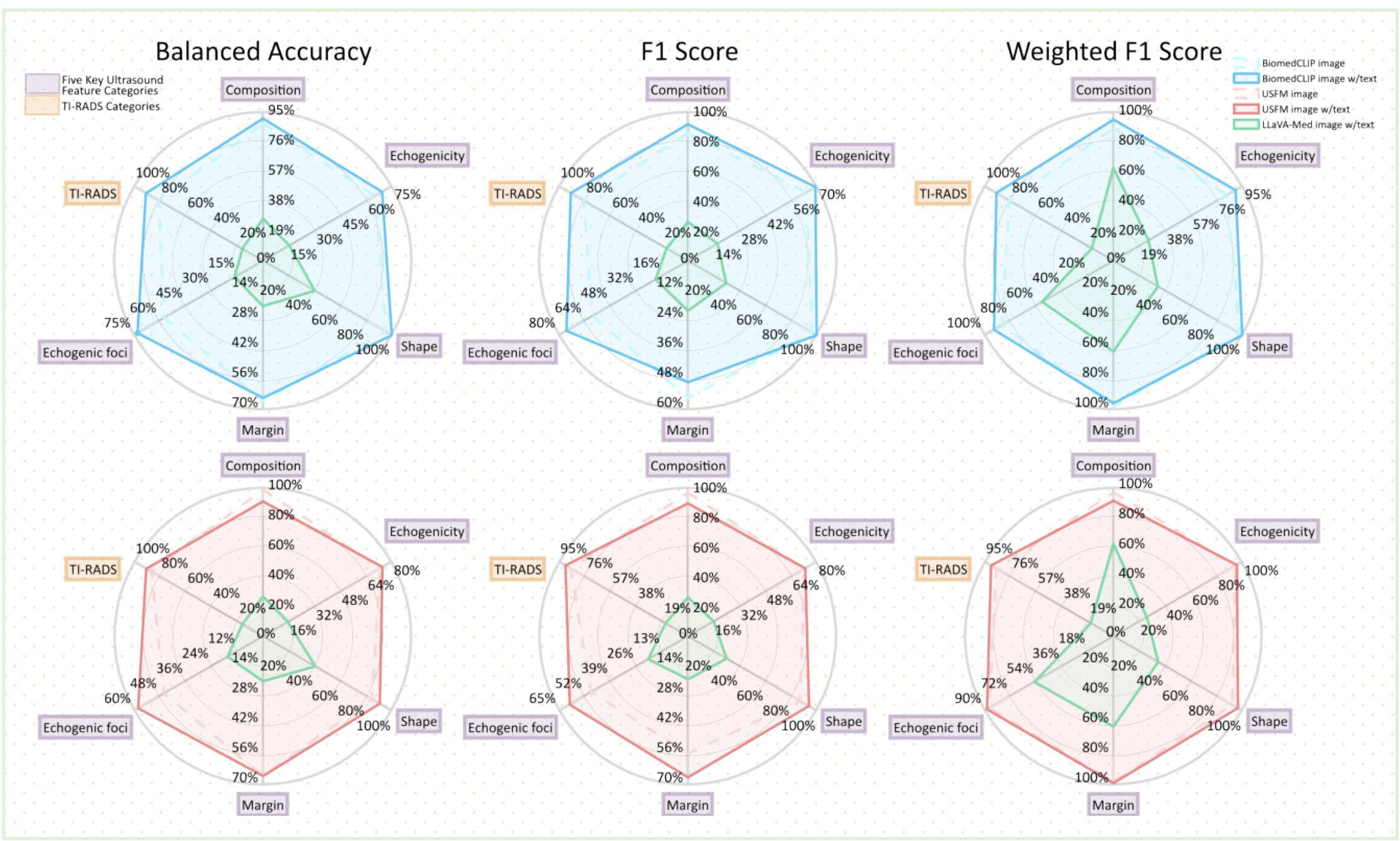

**Figure 8. Performance comparison of embedding alignment and direct image-only optimization.** Results compare CMCNet (image-text alignment) with FLaMM-Net (image-only multitask) and the VQA-style baseline LLaVA-Med, under two matched settings: BiomedCLIP with Multi-head v1 and USFM with Multi-head v2. Metrics include balanced accuracy, F1, and weighted F1. For CMCNet, BiomedCLIP uses L2-CMCL and USFM uses L1-CMCL.

We compare embedding alignment against direct optimization from images in this section. Using the same encoders and head variants, CMCNet improves over FLaMM-Net in most task–metric combinations (Figure 8). Under BiomedCLIP with Multi-head v1, CMCNet with L2-CMCL outperforms FLaMM-Net on five of six tasks across all three metrics, with margin as the main exception. Under USFM with Multi-head v2, CMCNet with L1-CMCL shows the same pattern, with composition as the main exception. Aggregated across tasks and encoders, these results indicate that aligning intermediate image embeddings to structured textual anchors is generally more effective than optimizing TI-RADS and attribute heads solely from images.

We further compare both discriminative baselines to the VQA-style model LLaVA-Med (Li C *et al.* 2023). LLaVA-Med is consistently below FLaMM-Net and CMCNet across metrics and encoders, which is expected under two practical constraints. First, publicly available medical VQA checkpoints often rely on a vision encoder trained primarily on natural images, which can limit transfer to ultrasound and reduce label efficiency in small datasets. Second, closed-set attribute prediction under strong class imbalance can amplify shortcut learning and language priors in VQA-style training, a known issue in the VQA literature (Goyal *et al.* 2017, Hudson and Manning 2019). For this setting, discriminative training with explicit embedding alignment provides a more direct inductive bias than generative VQA.

# 5 Conclusion

This study addresses the absence of structured feature-level supervision in thyroid ultrasound datasets by constructing the STN dataset, which contains 600 single-nodule cases with transverse and longitudinal views, bounding box annotations, and expert-labeled TI-RADS attributes. We first demonstrate that TI-RADS categories can be accurately predicted from textual representations of the five key ultrasound features using a simple linear head. This data confirms that structured feature descriptions encode sufficient diagnostic information. However, directly relying on text embeddings is not clinically meaningful at inference time. This observation motivates a training paradigm that leverages feature-level supervision and textual structure during training while remaining image-only at deployment.

We compare two strategies for ultrasound-based TI-RADS classification: direct multi-task optimization from images (FLaMM-Net) and intermediate embedding alignment between images and frozen textual representations (CMCNet). Experimental results consistently show that embedding alignment yields superior performance across attribute-level and TI-RADS prediction tasks. The improvement is particularly evident in fine-grained features such as margin and echogenic foci. These findings indicate that direct optimization from images can yield multiple observationally plausible yet clinically inconsistent solutions. In contrast, alignment to structured textual anchors constrains the embedding space toward a decision function that more faithfully reflects the TI-RADS scoring framework and enhances clinical interpretability.

To improve the mapping between image and text embeddings, we introduce Center-Margin Contrastive Loss (CMCLoss), which jointly enforces intra-class compactness and cross-modal alignment. CMCLoss outperforms InfoNCE and Center Loss across configurations, and provides a practical solution for structured image-text alignment. Comparison with the VQA-style multimodal baseline LLaVA-Med further indicates that general-purpose generative models are not well suited for small, imbalanced, closed-set ultrasound classification tasks without domain-specific pretraining. Overall, the results demonstrate that structured embedding alignment offers a more effective framework for ultrasound-based thyroid nodule classification than direct image-only optimization.

Future work will focus on expanding the STN dataset to include larger, multi-center cohorts and multi-nodule cases in order to improve generalization, address class imbalance, and enable external validation under device and institutional variability. Methodologically, more principled alignment objectives beyond CMCLoss should be investigated, and large-scale self-supervised or domain-adaptive pretraining tailored to thyroid ultrasound may further enhance robustness. With increased data scale, the framework can be extended toward a dedicated TI-RADS foundation model and a clinically oriented question–answer diagnostic system grounded in structured ultrasound feature representations.

# 6 Acknowledgements

This work was supported by the National Natural Science Foundation of China (No. 32201234), the Natural Science Foundation of Jilin Province YDZJ202301ZYTS401, and the Fundamental Research Funds for the Central Universities (JLU).

# Declaration of Statements

## Ethics Statement

This study was approved by the Institutional Review Board of China-Japan Union Hospital of Jilin University (Approval No. 2026030504). Owing to the retrospective design and the use of fully anonymized patient data, the requirement for written informed consent was waived.

## Author Contributions

B.Y. and X.W. were jointly responsible for the research concept and model design; B.Y. implemented the algorithm code and conducted the main experiments; J.Z., W.W., and L.W. participated in data preprocessing; L.H. and X.F. participated in experimental validation and results analysis; F.Z. and K.L. proposed the core research ideas, provided overall guidance, and made critical revisions to the manuscript. All authors reviewed and revised the manuscript and approved the final version for submission.

## Data Availability

The dataset is freely available at doi: 10.5281/zenodo.19125693 and the source code is available at: https://www.healthinformaticslab.org/supp/.

# 8 Appendix

## 8.1 Environmental and Experimental Settings

### 8.1.1 Training Protocol

All experiments were conducted on a single NVIDIA RTX 4070 Ti SUPER GPU with 16 GB memory. FLaMM-Net and CMCNet were trained using identical data splits to ensure fair comparison. The STN dataset was stratified by TI-RADS category and divided into training, validation, and test sets in a ratio of 10:1:1.

A linear learning rate warmup strategy was adopted, with warmup steps set to 5% of the total training iterations. For USFM, which is a pretrained ultrasound foundation model, a relatively small learning rate was required to preserve pretrained representations and maintain training stability. In contrast, BiomedCLIP was fine-tuned with a larger learning rate. The main hyperparameters are summarized in **Appendix Table A1**. Unless otherwise specified, optimization settings were kept consistent across configurations.

**Appendix Table A1. Hyperparameter settings for FLaMM-Net and CMCNet.**

| Architecture | Image Encoder | Learning Rate | Batch Size | Epochs |
|---|---|---|---|---|
| **FLaMM-Net** | USFM | 1.0e-5 | 4 | 50 |
| | BiomedCLIP | 1.0e-5 | 4 | 50 |
| | SAMUS | 1.0e-5 | 4 | 50 |
| | ResNet | 1.0e-3 | 4 | 50 |
| | DenseNet | 1.0e-3 | 4 | 50 |
| | EfficientNet | 1.0e-3 | 4 | 50 |
| **CMCNet** | USFM | 1.0e-5 | 128 | 200 |
| | BiomedCLIP | 1.0e-3 | 128 | 50 |

### 8.1.2 Inference Protocol

For model selection, the checkpoint achieving the highest TI-RADS classification accuracy on the validation set was used for final evaluation. All reported results are obtained from the held-out test set.

During inference, the model relies solely on ultrasound images. The five key ultrasound feature annotations and point-based scoring information are used only during training and are not required at test time.

## 8.2 Feature Annotation Dictionary

The following dictionary defines the categorical codes used for ultrasound feature annotations in the STN dataset (Appendix Table A2). These categories correspond to the five key ultrasound attributes in the TI-RADS scoring framework, along with nodule location metadata.

These categorical encodings are used for supervised training and evaluation. Location is provided as contextual metadata and is not included in the TI-RADS scoring calculation.

**Appendix Table A2. Ultrasound feature coding scheme.**

| Feature | Code | Description |
| --- | --- | --- |
| Composition | 0 | Cystic or spongiform |
| | 1 | Mixed cystic and solid |
| | 2 | Solid |
| Echogenicity | 0 | Anechoic |
| | 1 | Hyperechoic or isoechoic |
| | 2 | Hypoechoic |
| | 3 | Very hypoechoic |
| Shape | 0 | Wider-than-tall |
| | 3 | Taller-than-wide |
| Margin | 0 | Smooth |
| | 1 | Ill-defined |
| | 2 | Lobulated or irregular |
| | 3 | Extra-thyroidal extension |
| Echogenic Foci | 0 | None or large comet-tail artifacts |
| | 1 | Macrocalcifications |
| | 2 | Peripheral calcifications |
| | 3 | Punctate echogenic foci |
| Location | Left | Left lobe |
| | Right | Right lobe |
| | Middle | Isthmus |

## 8.3 Visualization of Image and Text Embeddings

We visualize the image embeddings learned by CMCNet using fine-tuned BiomedCLIP and USFM encoders, together with the text embeddings produced by the frozen ModernBERT encoder prior to projection. All embeddings are projected into a two-dimensional space using PCA for qualitative inspection. As shown in Appendix Figure A1, text embeddings form compact and well-concentrated clusters, reflecting their structured and low-variance semantic representation. In contrast, image embeddings exhibit a broader spatial distribution, indicating higher variability induced by visual appearance and acquisition factors.

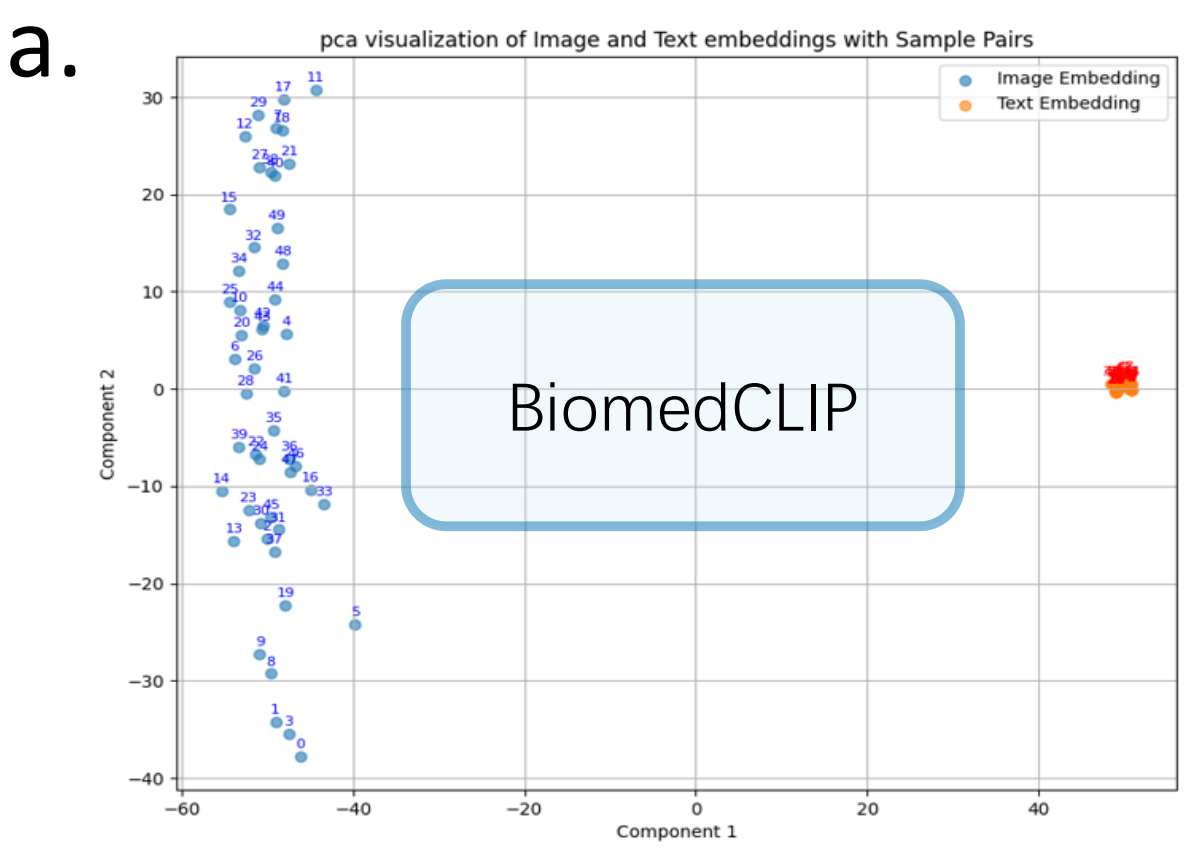

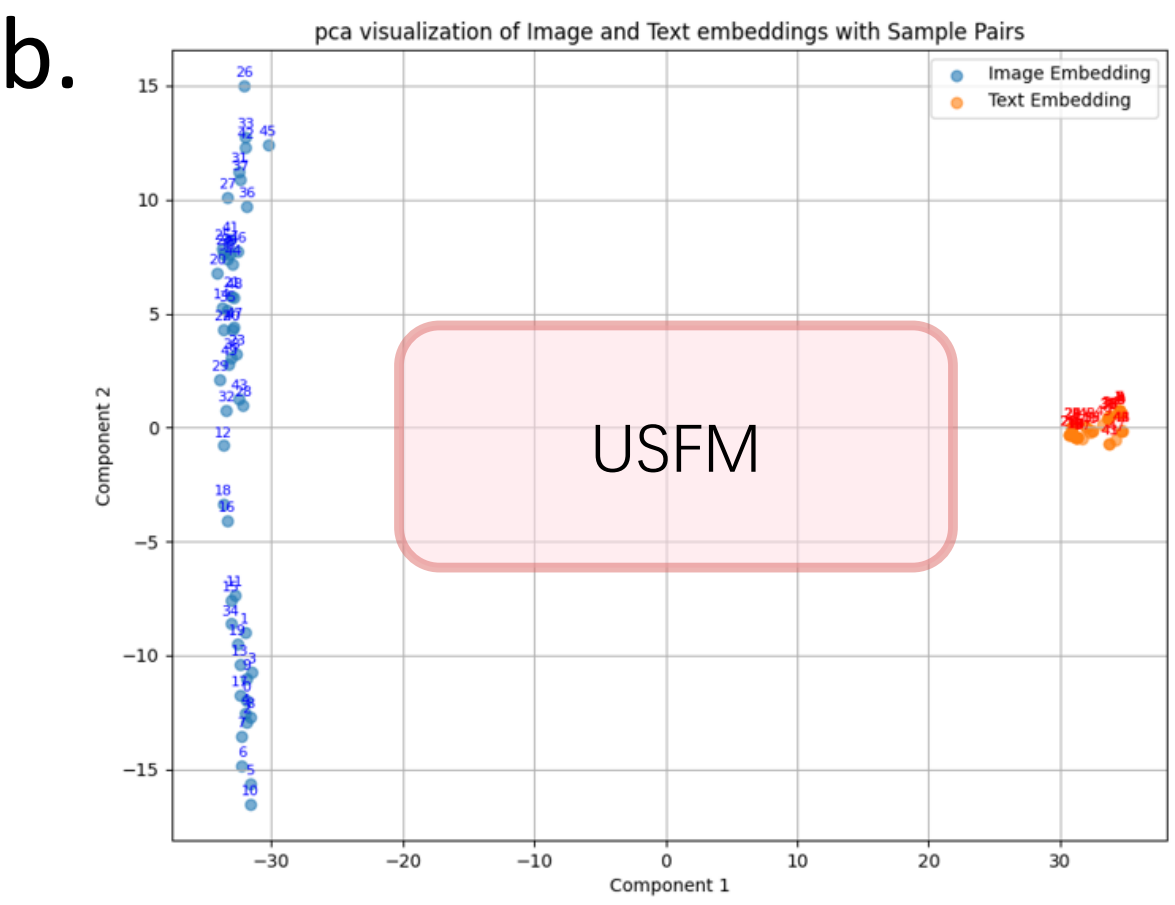

Appendix Figure A1. PCA visualization of image and text embeddings. The figure shows image embeddings learned by CMCNet with BiomedCLIP (a) and USFM (b), alongside text embeddings generated by the frozen ModernBERT encoder before projection. Text embeddings are tightly clustered, while image embeddings are more dispersed, highlighting the role of embedding alignment in reducing the modality gap.